%% file: main.tex
\documentclass{article}

\usepackage[numbers]{natbib}
\usepackage[preprint]{neurips_2024}

\usepackage[dvipsnames]{xcolor}         
\definecolor{linkColor}{rgb}{0.2,0.4,0.6}
\usepackage[utf8]{inputenc} 
\usepackage[T1]{fontenc}    
\usepackage[colorlinks=true,linkcolor=linkColor,citecolor=linkColor,filecolor=linkColor,urlcolor=linkColor]{hyperref}       
\usepackage{url}            
\usepackage{booktabs}       
\usepackage{amsfonts}       
\usepackage{nicefrac}       
\usepackage{microtype}      

\usepackage{graphicx}
\usepackage{arydshln}

\usepackage{booktabs}
\usepackage{multirow}
\usepackage{caption}
\usepackage{subcaption}
\usepackage{makecell}
\usepackage{csquotes}
\usepackage{epigraph}
\usepackage{relsize}

\RequirePackage{algorithm}
\RequirePackage{algorithmic}

\input{settings.tex}
\input{math_commands.tex}

\newcommand\our{LatentLM}

\newcommand\ourtok{$\sigma$-VAE}

\title{Multimodal Latent Language Modeling with Next-Token Diffusion}

\vspace{-1em}
\author{
Yutao Sun\thanks{~Core contributors.~~${\triangle}$ Tech lead.~~$\diamond$ Corresponding author.}$~~^{\dag\ddag}$~~~~Hangbo Bao\footnotemark[1]$~~^{\dag}$~~~~Wenhui Wang\footnotemark[1]$~~^{\dag}$~~~~Zhiliang Peng\footnotemark[1]$~~^{\dag}$~~~~Li Dong\footnotemark[1]$~~^{\triangle}$$^{\dag}$\\
\bf Shaohan Huang$^{\dag}$~~~Jianyong Wang$^{\ddag}$~~~Furu Wei$^{\dag}$$^{\diamond}$ \\
$^\dag$ Microsoft Research ~~~~
$^\ddag$ Tsinghua University \\
{\href{https://aka.ms/GeneralAI}{https://aka.ms/GeneralAI}}
}

\begin{document}

\maketitle
\vspace{-1em}
\begin{abstract}
Multimodal generative models require a unified approach to handle both discrete data (e.g., text and code) and continuous data (e.g., image, audio, video). In this work, we propose Latent Language Modeling (\our{}), which seamlessly integrates continuous and discrete data using causal Transformers. Specifically, we employ a variational autoencoder (VAE) to represent continuous data as latent vectors and introduce next-token diffusion for autoregressive generation of these vectors. Additionally, we develop $\sigma$-VAE to address the challenges of variance collapse, which is crucial for autoregressive modeling. Extensive experiments demonstrate the effectiveness of \our{} across various modalities. In image generation, \our{} surpasses Diffusion Transformers in both performance and scalability. When integrated into multimodal large language models, \our{} provides a general-purpose interface that unifies multimodal generation and understanding. Experimental results show that \our{} achieves favorable performance compared to Transfusion and vector quantized models in the setting of scaling up training tokens. In text-to-speech synthesis, \our{} outperforms the state-of-the-art VALL-E 2 model in speaker similarity and robustness, while requiring 10$\times$ fewer decoding steps. The results establish \our{} as a highly effective and scalable approach to advance large multimodal models.
\end{abstract}

\vspace{-0.4em}
\begin{figure}[h]
\centering
\includegraphics[width=0.9\linewidth]{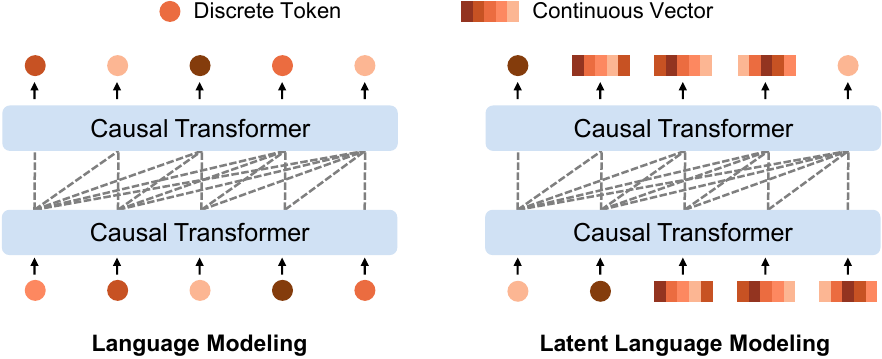}
\caption{Latent Language Modeling (\our{}) seamlessly handles continuous (e.g., image, audio, video) and discrete (e.g., text and code) data using causal Transformers. We introduce next-token diffusion to autoregressively generate the latent vectors one by one. 
The proposed method provides a general-purpose interface that unifies multimodal generation and understanding.
}
\label{fig:intro}
\end{figure}

\newpage
\section{Introduction}
\label{sec:intro}

Multimodal generative models need a unified modeling method to process both discrete data (e.g., text and code) and continuous data (e.g., video, audio, and robot actions).
Most previous systems rely on building pipelines or calling external tools. For example, language models perceive and produce audio or image data using independent modules, i.e., automatic speech recognition, text-to-speech, and text-to-image models.
However, it is difficult to perform end-to-end optimization for pipeline-based methods. Information loss between modules also restricts performance, as the modules typically use text prompts for communication.

In order to natively handle discrete and continuous data in multimodal large language models, there have been three main strands of research.
The first one~\cite{dalle,valle,chameleon} uses VQ-VAE~\cite{vqvae,vqgan} to quantize continuous data into discrete codes and treats everything as discrete tokens in autoregressive language models.
The continuous data are then recovered by the VQ-VAE decoder by conditioning on discrete codes.
The performance is often limited by lossy tokenization, which creates a restrictive bottleneck during quantization.
The low compression ratio also renders the length of discrete codes long.
Given the success of diffusion models on continuous data generation~\cite{ddpm,sd}, another strand of work~\cite{one:trm:fit,codi} unifies the modeling of discrete data into diffusion models.
However, the unification compromises by following the diffusion-based method, which harms the modeling performance of discrete data.
The third strand of research~\cite{transfusion} shares model weights while using sequence-level diffusion for continuous data and next-token prediction for discrete data.
Although sharing parameters, they have different objectives (i.e., denoising for diffusion of continuous data and next-token prediction for discrete data) and implementation details (i.e., bidirectional attention for diffusion and causal attention for next-token prediction).
The bidirectional diffusion also restricts the model's applications to variable-length sequences.
Moreover, the noise added in diffusion training interferes with joint training of interleaved data.

In this work, we propose latent language modeling (\our{}), which seamlessly supports continuous and discrete data with causal Transformers in a unified manner.
Specifically, we represent continuous data as latent vectors using variational autoencoder (VAE).
We introduce next-token diffusion to autoregressively predict the latent vectors, where diffusion heads produce latent vectors by conditioning on each Transformer hidden state.
Then the generated continuous data can be recovered by the VAE decoder.
For discrete data, the shared Transformer backbone is used to perform next-token prediction with softmax heads.
Moreover, in order to make representations suitable for autoregressive decoding, we propose \ourtok{} to maintain the variance of the latent space.

\our{} unifies the generation of discrete and continuous tokens under the language modeling paradigm, allowing information sharing among different modalities.
The proposed method simplifies implementation by reusing the existing distributed training infrastructure of large language models.
Another advantage is that \our{} unifies generation and understanding with a general-purpose interface, which perceives and produces any combination of multimodal data, e.g., text, image, audio, video, and robot action data.
Compared to quantizing continuous data, \our{} has a higher compression ratio while maintaining relatively lossless reconstruction quality.

We conduct experiments on image generation, multimodal large language models, and text-to-speech synthesis to show the flexibility and effectiveness of \our{} across modalities.
First, image generation on ImageNet~\cite{imagenet} shows that \our{} achieves competitive performance with the models based on diffusion or discrete tokens.
The results demonstrate that \our{} outperforms DiT~\cite{dit} in the setting of scaling model size.
Second, we train multimodal large language models with text, image-text pairs, and interleaved data.
The results show that \our{} outperforms Transfusion~\cite{transfusion} and the model with vector-quantized image tokenizers, in terms of language modeling, text-to-image generation, and vision-language understanding metrics.
We also scale up the number of training tokens and find that \our{} has favorable scaling properties.
Third, experimental results on text-to-speech synthesis show that \our{} achieves better performance than previous systems.
Because our tokenizer uses continuous representations, the compression ratio is much larger than previous vector-quantized tokenizers, which improves both the training and inference efficiency.

\vspace{2.8em}

\newpage
\section{Latent Language Modeling}
\label{sec:methods}

\begin{figure}[t]
\centering
\includegraphics[width=0.9\linewidth]{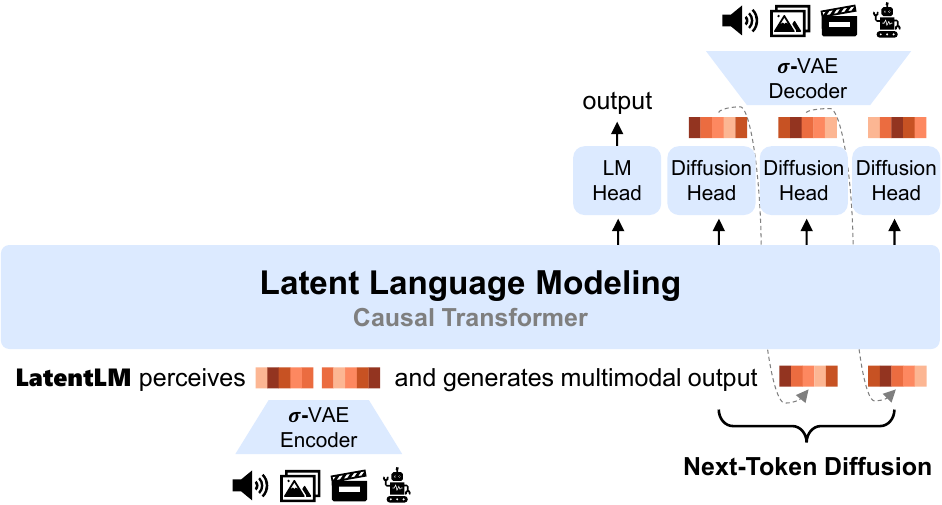}
\caption{\our{} unifies the modeling of continuous and discrete data. We introduce $\sigma$-VAE (\Cref{sec:vae}) to represent continuous data as latent vectors. We perform next-token diffusion (\Cref{sec:ntd}) to autoregressively predict the latent vectors one by one. The diffusion head generates vectors by conditioning on the output states of Transformer. The predicted vectors can be decoded to produce the final outputs.}
\label{fig:method}
\end{figure}

Latent language modeling (\our{}) autoregressively perceives and generates multimodal sequences (with discrete and continuous data) in a unified way.
As shown in \Cref{fig:method}, the model is a causal Transformer, where the $t$-th token is predicted by conditioning on previous $t-1$ tokens.
Continuous data are generated by next-token diffusion (\Cref{sec:ntd}), where the diffusion head is used to produce continuous vectors for each position.
In addition, discrete tokens are generated by next-token prediction, similar to conventional language modeling.

Specifically, let $x = x_1 \cdots x_{N}$ denote an input sequence of discrete and continuous tokens.
For a discrete token, we use a lookup table to get its vector representation.
For continuous data, variational autoencoder (VAE)~\cite{vae} is used as tokenizer to compress input data to latent vectors (\Cref{sec:vae}).
After obtaining the vector representations, we pack the input vectors into $X^0 = [\vx_1, \cdots, \vx_{N}] \in \mathbb{R}^{N\times d}$, where $d$ represents the hidden dimension of the model.
$X^0$ is fed into a language model based on causal Transformer.

The language model is stacked with $L$ Transformer layers.
Causal masking is used for autoregressive generation.
We also adopt pre-RMSNorm~\cite{rmsnorm} and SwiGLU~\cite{glu,swish} as improvements after LLaMA~\cite{llama}.
The input $X^0$ is further contextualized to obtain the output $X^L$, i.e., $X^{l} = \operatorname{Decoder}(X^{l-1}), \ l \in [1, L]$.
The output states of Transformer $[\vh_1, \cdots , \vh_N] = \mathrm{RMSNorm}(X^L)$ are used to decode the predictions:
\begin{equation}
\begin{aligned}
\mathrm{Decode}({x}_i | {x}_{<i}) &= \left\{\begin{array}{ll}
\mathrm{Sample}\left( P_d({x}_i | {x}_{<i}) \right) & \text{$x_i$ is a discrete token} \\
\mathrm{Diffusion}( \vh_i ) & \text{$x_i$ is a continuous vector}
\end{array}\right. \\
P_d({x}_i | {x}_{<i}) &= \softmax(\vh_i W_v)
\label{eq:lm:decode}
\end{aligned}
\end{equation}
where $W_v \in \mathbb{R}^{d \times |\mathcal{V}|}$ is the $\softmax$ classifier weight, $|\mathcal{V}|$ is the vocabulary size, and $\mathrm{Sample}(\cdot)$ is a sampling algorithm (e.g., greedy decoding, and top-$p$ sampling).
The $\mathrm{Diffusion}(\cdot)$ head is described in \Cref{sec:ntd}, which decodes continuous vectors by conditioning on the hidden state $\vh_i$.
The latent vectors are generated autoregressively one by one, i.e., next-token diffusion.
Then the VAE decoder is used to generate raw data from the predicted latent vectors.

\subsection{Next-Token Diffusion}
\label{sec:ntd}

\our{} autoregressively generates the continuous tokens.
We use diffusion as the language model head for each continuous token.
The diffusion head progressively refines and generates the latent vector $\vx_i$ by conditioning on the hidden state $\vh_i$.
Then the predicted $\vx_i$ is used as input for the next step of Transformer.

In our experiments, we use either denoising diffusion probabilistic model (DDPM) \cite{ddpm} or flow matching~\cite{flow} as our design choice.
We use DDPM as an example to describe the details.
Diffusion is formulated as two processes, i.e., the forward process gradually adds noise to the input, and the reverse process learns to denoise step by step.

\paragraph{Forward Process}
Noise is introduced incrementally into the original vector in $T$ steps.
Let ${\vx}_{i}^{0} = {\vx}_{i}$ denote the original data and ${\vx}_{i}^{t}$ the noisy version, where $t=1,\cdots,T$.
The Markov noise-addition process is defined as $q({\vx}_{i}^{t}|{\vx}_{i}^{t-1}) = \mathcal{N}({\vx}_{i}^{t} ; \sqrt{1-{\beta}_t} {\vx}_{i}^{t-1}, \beta_t \mI)$, where Gaussian noise is injected in each step, $\beta_t$ follows a predefined noise schedule, and $\mI$ is the identity covariance matrix.
A nice property is that we can directly sample ${\vx}_{i}^{t}$ from the original data ${\vx}_{i}$ through:
\begin{equation}
{\vx}_{i}^{t} = \sqrt{\overline{\alpha}_t}{\vx}_{i} + \sqrt{1-\overline{\alpha}_t} \boldsymbol{\epsilon}
\label{eq:ddpm:forward}
\end{equation}
where $\overline{\alpha}_t = \prod_{i=1}^{t}(1-\beta_i)$, and $\boldsymbol{\epsilon} \sim \mathcal{N}(0,\mI)$.

\paragraph{Reverse Process}
The diffusion head is trained to denoise the data step by step to recover the original vectors, which is parameterized by a probabilistic model $p_\theta({\vx}_{i}^{t-1}|{\vx}_{i}^{t}, \vh_i)$.
DDPM learns a model $\epsilon_\theta({\vx}_{i}^{t}, t, \vh_i)$ to estimate the noise $\boldsymbol{\epsilon}$ (as described in \Eqref{eq:ddpm:forward}) of ${\vx}_{i}^{t}$ in the $t$-th step, conditioning on the Transformer state $\vh_i$.
The model parameters are learned by minimizing the following loss:
\begin{equation}
\mathcal{L}_{\text{Diff}}({\vx}_i , \vh_i) = \mathbb{E}_{\vx_{i}, t, \boldsymbol{\epsilon}} \parallel \boldsymbol{\epsilon} - \boldsymbol{\epsilon}_\theta({\vx}_{i}^{t}, t, \vh_i) \parallel^2
\label{eq:ddpm:reverse}
\end{equation}
where $\boldsymbol{\epsilon}$ is the actual Gaussian noise.

\paragraph{Head Architecture}
We use a lightweight neural network as $\boldsymbol{\epsilon}_{\theta}(\cdot)$ in \Eqref{eq:ddpm:reverse}, which is a residual architecture incorporating pre-RMSNorm~\cite{rmsnorm} and feedforward networks~\cite{mar}.
The network input is a vector that contains noise.
The output is the predicted noise $\boldsymbol{\epsilon}_{\theta}(\cdot)$.
We also utilize AdaLN-Zero~\cite{dit} which conditions on both the timestep $t$ and the Transformer output $\vh_i$.
This head processes a noised continuous vector and predicts the corresponding noise.

\paragraph{Inference}
The Transformer state $\vh_i$ is used as the condition for diffusion head.
The diffusion process iteratively denoises data.
At first, a vector of pure Gaussian noise $\vx_T$ is given.
In each step, the predicted noise $\boldsymbol{\epsilon}_\theta({\vx}_{i}^{t}, t, \vh_i)$ is used to produce $\vx_{t-1}$ from $\vx_t$, which also considers the noise schedule for scaling~\cite{ddpm}.
In our experiments, we utilize DPM-Solver~\cite{dpm-solver,dpm-solver++} to accelerate the denoising process, significantly reducing the number of inference steps compared to the training phase.

\subsection{Model Training and Inference}

\paragraph{Training}
During training, we compute the token-level loss for training sequences.
For discrete data, we use the standard language modeling objective to maximize the likelihood of data.
Specifically, the loss is computed as $\mathcal{L}_{\text{LM}} = -\sum_{x,i}{\log P_d({x}_i | {x}_{<i})}$, where the prediction probability is presented in \Eqref{eq:lm:decode}.
For continuous data, the loss function $\mathcal{L}_{\text{Diff}}$ described in \Eqref{eq:ddpm:reverse} is used.
The training objective is to minimize ${\mathcal{L}_{\text{LM}} + \alpha \mathcal{L}_{\text{Diff}}}$, where $\alpha$ is a hyperparameter.
In practice, we sample multiple diffusion timesteps, typically four, for a single forward pass~\cite{mar}.
As the diffusion head is usually lightweight, reusing the computation of the Transformer backbone improves training efficiency while introducing minimal overhead.

\paragraph{Inference}
The decoding process is similar to that of standard causal Transformers, i.e., predicting the next token based on the generation history that has come before it.
The tokens are produced following \Eqref{eq:lm:decode}.
Notice that the Transformer backbone is computed in a single pass, and only the lightweight diffusion head requires multiple denoising steps.
In addition, we use special tokens to indicate the switch between the language modeling head and the diffusion head.
For instance, we use \texttt{<BOD>} to denote the beginning of the diffusion head usage, and \texttt{<EOD>} to indicate the switch back to the language modeling head.

\subsection{Latent Vector Representation of Continuous Data}
\label{sec:vae}

\begin{figure}[t]
\centering
\includegraphics[width=\linewidth]{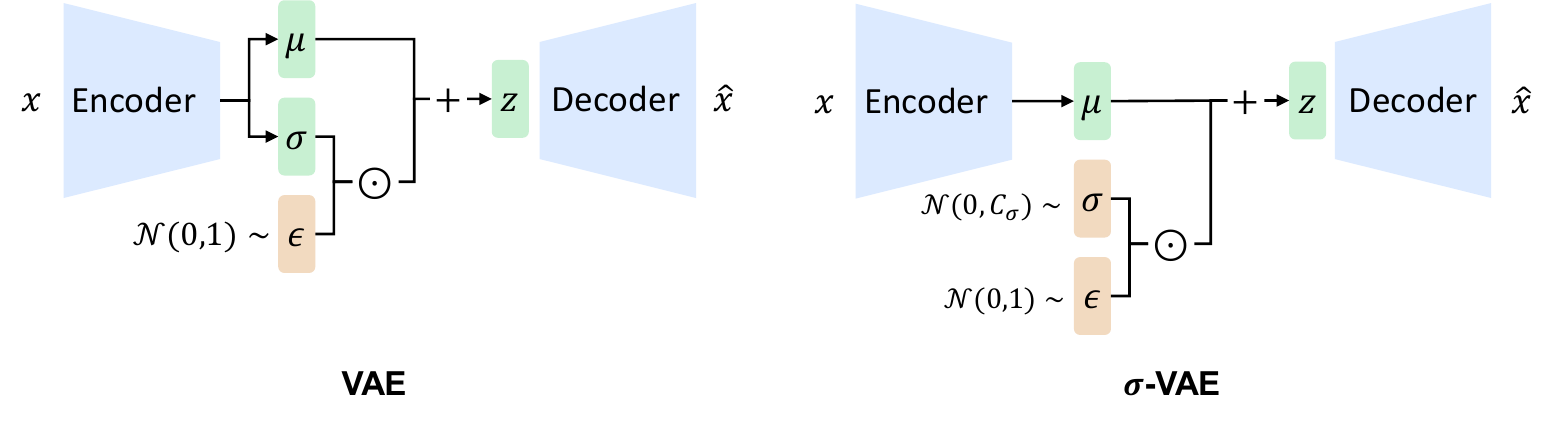}
\caption{Compared to variational autoencoder (VAE), \ourtok{} uses a fixed variance for the latent space. It avoids variance collapse and makes \our{} more robust to exposure bias during autoregressive generation. In our method, $\sigma$ is a scalar that is sampled from $\mathcal{N}(0, C_{\sigma})$ for each example.}
\label{fig:vae}
\end{figure}

The tokenizer compresses continuous data into latent vectors.
It is based on variational autoencoder (VAE)~\cite{vae}, which encodes the input data into a latent space and then decodes it back to the original space.
Let $x$ denote the continuous input and $z$ the compressed vector representations.
VAEs maximize the evidence lower bound of log-likelihood $\log p(x)$ via:
\begin{equation}
\mathrm{maximize}~~\mathbb{E}_{q_{\phi}(z|x)} \left[ \log p_{\psi}(x|z) \right] - \mathcal{D}_{\text{KL}} \left[ q_{\phi}(z|x) \parallel p(z) \right]
\label{eq:l_vae}
\end{equation}
where the encoder $q_{\phi}(z|x)$ encodes input $x$ to latent vectors $z$, the decoder $p_{\psi}(x|z)$ reconstructs data by conditioning on $z$, and the KL term encourages that the latent space follows a Gaussian prior distribution.

Because autoregressive generation introduces sampling uncertainty, the representation variance of the latent space affects the performance of next-token diffusion.
Larger variance of latent representation makes the model more robust to exposure bias during inference~\cite{givt}, as confirmed in \Cref{fig:var_fid}.
However, for vanilla VAEs, the variance of some channels tends to collapse, which harms autoregressive modeling.

In this work, we propose \ourtok{} to prevent variance collapse by enforcing a fixed variance in the latent space.
The reconstruction pass is computed as:
\begin{equation}
\begin{aligned}
\mu &=\mathrm{Encoder}_{\phi}(x) \\
z &= \mu + \sigma \odot \boldsymbol{\epsilon}, \text{where} ~ \boldsymbol{\epsilon} \sim \mathcal{N}(0, 1), ~ \sigma \sim \mathcal{N}(0, C_{\sigma}) \\
\hat{x} &= \mathrm{Decoder}_{\psi}(z)
\end{aligned}
\end{equation}
where $\sigma$ is a scalar, $C_{\sigma}$ is a hyperparameter, $\mathrm{Encoder}_{\phi}(\cdot)$ and $\mathrm{Decoder}_{\psi}(\cdot)$ are learnable models.
The input $x$ is fed into the encoder to obtain $\mu$.
The re-parameterization trick is used to make $z$ follow the Gaussian distribution.
The variance $\sigma$ is fixed across channels, and is sampled from $\mathcal{N}(0, C_{\sigma})$ for each example.
It allows us to manipulate the latent space to better align with the expectation of autoregressive models.
Then $z$ is fed into the decoder for reconstruction.
According to \Eqref{eq:l_vae}, the training objective of \ourtok{} is:
\begin{equation}
\mathrm{minimize}~~ \left\| \hat{x} - x \right\|_2^2 + \beta \left\| \mu \right\|_2^2
\end{equation}
where the first term is the reconstruction error, and the hyperparameter $\beta$ controls the trade-off between reconstruction quality and adherence to the prior distribution~\cite{beta-vae}.

\section{Experiments}
\label{sec:exp}

We evaluate \our{} through multiple dimensions to thoroughly assess its effectiveness and scalability.
We conduct experiments on various types of tasks and modalities as follows:
\begin{itemize}[nosep,leftmargin=*]
\item \Cref{sec:exp:image}: Image Generation\footnote{~The code and pretrained weights are available at \url{https://aka.ms/next-token-diffusion}.}
\begin{itemize}
\item Category $\rightarrow$ Image
\end{itemize}
\item \Cref{sec:exp:mllm}: Multimodal Large Language Models
\begin{itemize}
\item 1) Interleaved Image-Text Data; 2) Text $\rightarrow$ Image; 3) Image $\rightarrow$ Text; 4) Text
\end{itemize}
\item \Cref{sec:exp:tts}: Text-to-Speech Synthesis
\begin{itemize}
\item Speech Prompt + Text $\rightarrow$ Speech
\end{itemize}
\end{itemize}



\subsection{Image Generation: Scalable Autoregressive Modeling}
\label{sec:exp:image}

The image generation experiments are conducted on ImageNet~\cite{imagenet}.
Given a category, the goal is to generate the corresponding images.
First, we systematically benchmark our model against state-of-the-art baselines to demonstrate the advantages of next-token diffusion.
We also investigate the scalability of our approach by evaluating it with larger model sizes and higher resolutions.
Furthermore, we compare the design choices of \ourtok{} tokenizers.
Finally, we assess the inference efficiency to highlight the practical deployment benefits of our method.

\subsubsection{System Evaluation}
\label{sec:exp:sys}

\begin{table}[t]
\centering
\begin{tabular}{l llc|cc}
\toprule
\bf Type & \bf Model & \bf \#Params & \bf \#Epochs & \bf FID$\downarrow$ & \bf IS$\uparrow$  \\
\midrule
\multicolumn{6}{l}{~~\textit{Non-Causal-Masking Generation}} \\
\multirow{3}{*}{\centering Diffusion} & LDM-4~\cite{sd} & 400M & --- &  3.60 & 247.7 \\
& DiT-XL/2~\cite{dit} & 675M & 400 & 2.27 & 278.2 \\
& U-ViT-H/2~\cite{u-vit} & 501M & 400 & 2.29 & 263.9 \\
\midrule
\multirow{2}{*}{\centering Masked Generative} & MaskGIT~\cite{maskgit} & 227M & 300 & 4.02 & 355.6 \\
& MAR-L~\cite{mar} & 479M & 800 & 1.78 & 296.0 \\
\midrule
\multicolumn{6}{l}{~~\textit{Causal-Masking Generation}} \\
\multirow{4}{*}{\centering Causal-Discrete} & VQGAN~\cite{vqgan} & 1.4B & 240 & 5.20 & 280.3 \\
& ViT-VQGAN~\cite{vit-vqgan} & 1.7B & 240 & 3.04 & 227.4 \\
& LlamaGen-XL~\cite{llamagen} & 775M & 300 & 2.62 & 244.1 \\
& LlamaGen-XXL~\cite{llamagen} & 1.4B & 300 & 2.34 & 253.9 \\
\midrule
\multirow{2}{*}{\centering Causal-Continuous} & GIVT-Causal-L+A~\cite{givt} & 1.67B & 500 & 2.59 & --- \\
& \our{}-L (This Work) & 479M & 400 & 2.24 & 253.8 \\
\bottomrule
\multicolumn{5}{c}{}
\end{tabular}
\caption{Image generation results on ImageNet~\cite{imagenet}. We evaluate FID~\cite{fid} and IS~\cite{metric:is}. \our{} achieves competitive performance, especially compared with other causal-masking image generation models.}
\label{tbl:imagenet_fid}
\end{table}

\paragraph{Setup}
We scale up model size and number of training steps.
We set the Transformer's hidden size to $1024$ and the number of layers to $32$.
The intermediate dimension of feedforward networks is $2730$.
The diffusion head has six layers.
We use the AdamW~\cite{adamw} optimizer with $\beta=(0.9,0.98)$.
We use a cosine learning rate schedule with the maximal value of 5e-4 and 100 warmup steps.
The weight decay is set to $0.1$.
We train models with 250,000 steps with batch size of 2048. The number of training epochs is about $400$.
Classifier-free guidance~\cite{cfg} is set to $1.65$.
As shown in \Cref{tbl:imagenet_fid}, the model configurations have been aligned with those of previous models to ensure fair comparisons.

\Cref{tbl:imagenet_fid} presents a comprehensive comparison between \our{} and various image generation methods.
These methods can be categorized into two main groups:
(1)~non-causal-masking models, including image-level diffusion models (LDM~\cite{sd}, DiT~\cite{dit}, U-ViT~\cite{u-vit}) and masked generative models (MaskGIT~\cite{maskgit}, MAR~\cite{mar}); and 
(2)~causal-masking models, comprising discrete-token generation approaches (VQGAN~\cite{vqgan}, ViT-VQGAN~\cite{vit-vqgan}, LlamaGen~\cite{llamagen}) and continuous autoregressive generation methods (GIVT-Causal~\cite{givt}).

\paragraph{Results}
\Cref{tbl:imagenet_fid} shows that \our{} achieves competitive performance compared to previous work.
Notice that non-causal-masking models typically require iterative forward computation during inference.
Consequently, the inference FLOPs of non-causal-masking models tend to be larger due to multiple forward passes.
Moreover, models using continuous representations typically outperform those using discrete code, even though \our{}-L has fewer parameters.
Among the methods, MAR~\cite{mar} and GIVT~\cite{givt} are the most relevant.
In comparison, MAR uses a bidirectional Transformer to implement masked autoregressive modeling, instead of causal Transformer, which renders MAR unable to reuse key-value caches for multiple forward passes.
Furthermore, unifying MAR and language modeling in multimodal models remains challenging due to their distinct modeling approaches.
In contrast, \Cref{sec:exp:mllm} shows that our approach can be naturally applied to multimodal large language models.
In addition, GIVT directly predicts latent vectors of VAEs with Gaussian mixture models.
The main difference is that \our{} integrates diffusion into causal Transformers, which tends to offer more powerful modeling expressivity.
The results also indicate that our approach outperforms GIVT with a smaller model size and fewer training epochs.

\subsubsection{Scalability}
\label{sec:exp:scaling}

\begin{wrapfigure}{r}{0.4\textwidth}
\setlength\intextsep{0pt}
\centering
\vspace{-1.5em}
\captionsetup{type=figure}
\includegraphics[width=\linewidth]{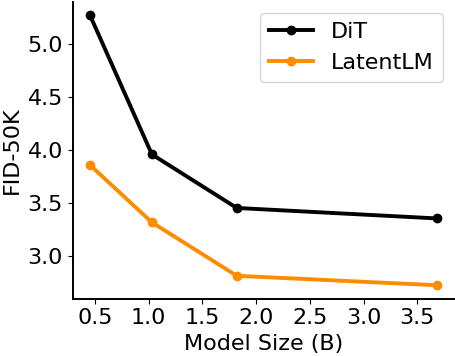}
\caption{Scaling curves of DiT and \our{}. FID~\cite{fid} consistently becomes better with larger model size.}
\label{fig:image:scaling}
\vspace{-1.5em}
\end{wrapfigure}

We compare the scalability properties of Diffusion Transformer (DiT)~\cite{dit} and \our{}, in terms of model size, and image resolution.

\paragraph{Setup}
In order to be consistent with \our{}, we also augment DiT with RMSNorm~\cite{rmsnorm} and SwiGLU~\cite{swish,glu}.
All models were trained with 75,000 steps, i.e., approximately $120$ epochs, for scaling experiments.
Classifier-free guidance~\cite{cfg} is set to $1.75$ during inference.
Detailed hyperparameters are presented in \Cref{app:hp:scaling}.

\paragraph{Scaling Model Size}
\label{sec:imagenet:scaling}
As shown in Figure~\ref{fig:image:scaling}, we trained models of varying sizes, i.e., 455M, 1.03B, 1.82B, 3.68B.
\our{} consistently outperforms DiT models.
The results demonstrate our approach's effective scaling properties in terms of model size.

\begin{figure}[t]
\centering
\begin{minipage}{0.24\textwidth}
\centering
\includegraphics[width=\linewidth]{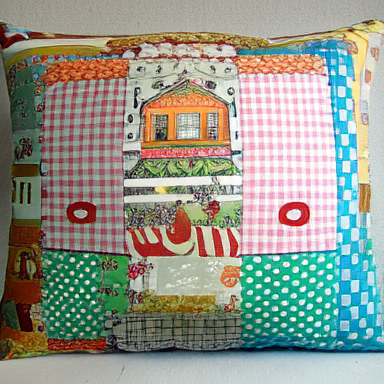}
\end{minipage}%
\begin{minipage}{0.24\textwidth}
\centering
\includegraphics[width=\linewidth]{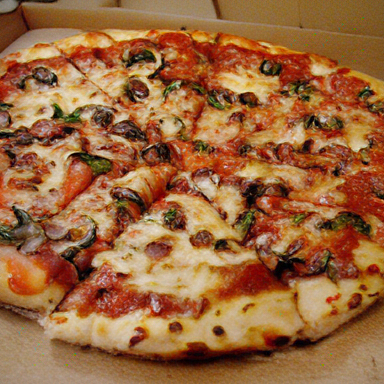}
\end{minipage}%
\begin{minipage}{0.24\textwidth}
\centering
\includegraphics[width=\linewidth]{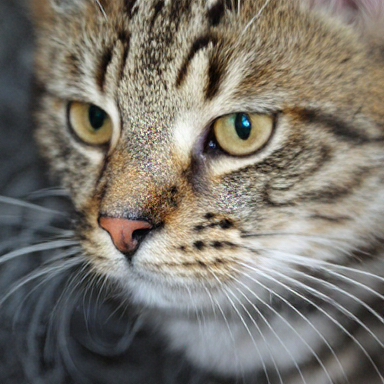}
\end{minipage}%
\begin{minipage}{0.24\textwidth}
\centering
\includegraphics[width=\linewidth]{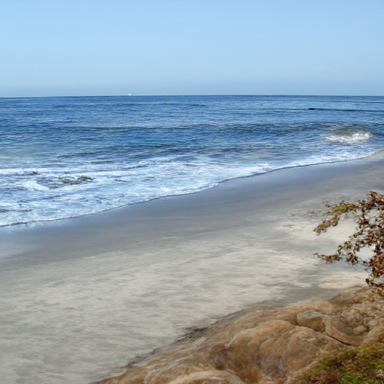}
\end{minipage}\\[-0.1em]  
\begin{minipage}{0.24\textwidth}
\centering
\includegraphics[width=\linewidth]{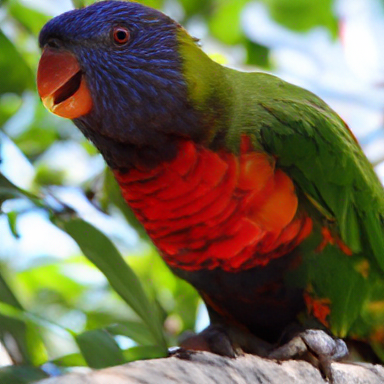}
\end{minipage}%
\begin{minipage}{0.24\textwidth}
\centering
\includegraphics[width=\linewidth]{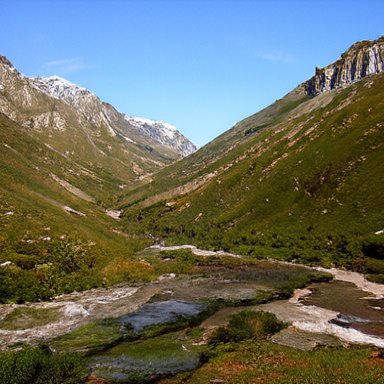}
\end{minipage}%
\begin{minipage}{0.24\textwidth}
\centering
\includegraphics[width=\linewidth]{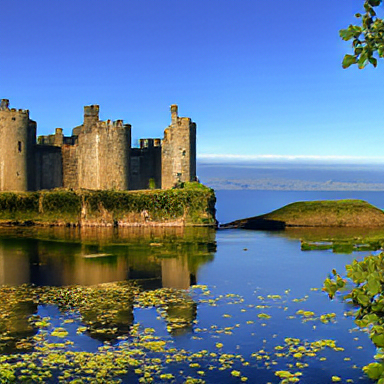}
\end{minipage}%
\begin{minipage}{0.24\textwidth}
\centering
\includegraphics[width=\linewidth]{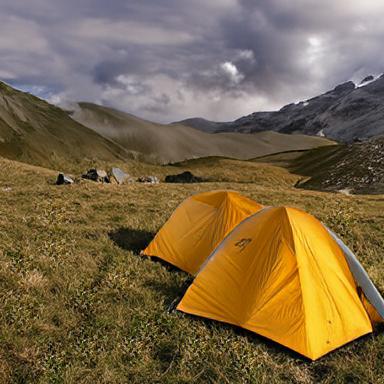}
\end{minipage}
\caption{Samples of \our{} trained on ImageNet. The resolution is 384$\times$384. The image is generated by models described in Section~\ref{sec:imagenet:scaling}.}
\label{fig:imagenet_samples}
\end{figure}

\begin{wraptable}{r}{0.28\textwidth}
\centering
\vspace{-2em}
\resizebox{\linewidth}{!}{%
\begin{tabular}{lc}
\toprule
\bf Resolution & \bf FID-50k$\downarrow$ \\
\midrule
$256\times256$ & 3.19 \\
$384\times384$ & \textbf{2.51} \\
\bottomrule
\\
\end{tabular}
}
\vspace{-1em}
\caption{FID~\cite{fid} of scaling up image resolution.}
\label{tbl:384}
\end{wraptable}

\paragraph{Scaling Resolution}
As shown in \Cref{tbl:384}, we conduct experiments at a resolution of 384, training a 1.82B model for 100,000 steps.
The results reveal significant improvements over the 256-pixel resolution when using classifier-free guidance~\cite{cfg}.
The improvement stems from the richer details and additional information captured in the tokenizer with higher resolutions.
Moreover, increasing resolution leads to longer sequences, which scales the decoding computation up.

\subsubsection{Effects of Tokenizer}
\label{sec:exp:tok}

\begin{figure*}
\centering
\includegraphics[width=\linewidth]{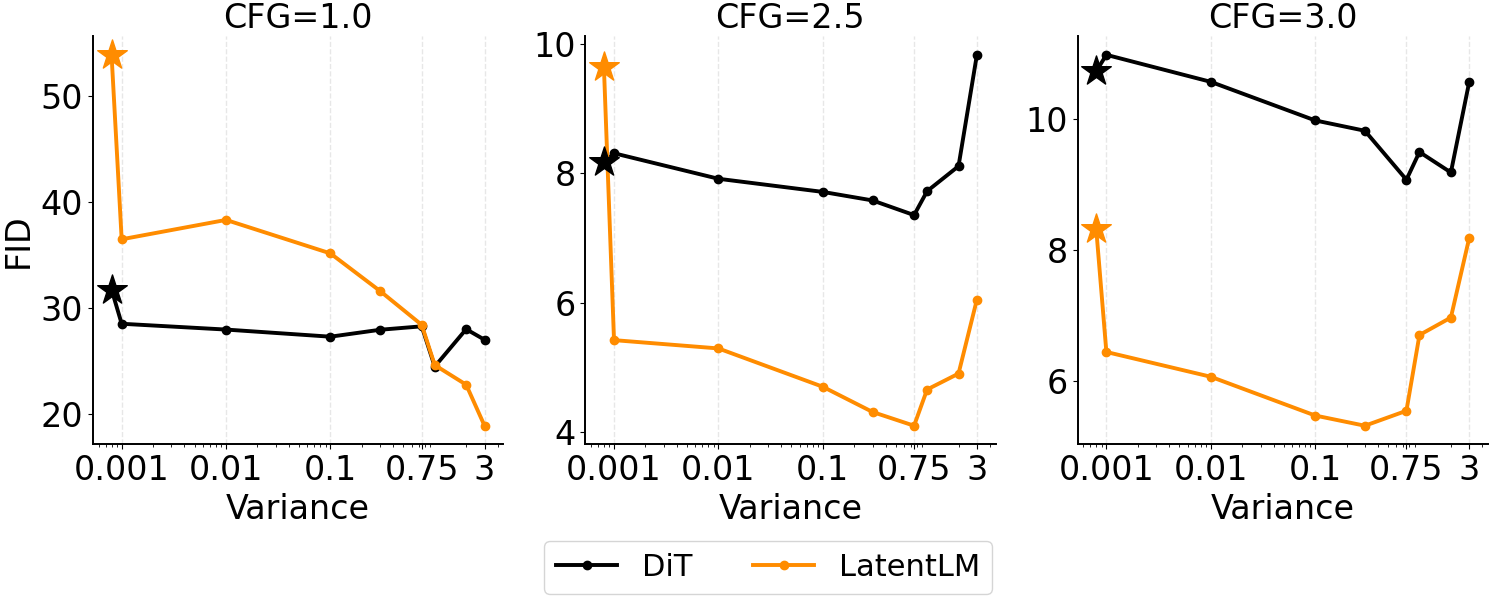}
\caption{Image generation results of Diffusion Transformer (DiT)~\cite{dit} and \our{} on ImageNet. We report FID~\cite{fid} scores (lower is better) in the settings of different tokenizer variance and CFG~\cite{cfg} scale.
The ``stars'' represent the tokenizers tuned for previous image-level diffusion models~\cite{sd}, which are ineffective for \our{}. The results indicate that \our{} favors tokenizers with larger variances.
}
\label{fig:var_fid}
\end{figure*}

As shown in \Cref{fig:var_fid}, we analyze the effects of \ourtok{} tokenizers with various configurations.
We evaluate their performance in both the DiT and \our{} frameworks.
Specifically, we train the \ourtok{} tokenizers with different variance.
To simplify the analysis, we use fixed variance values $\sigma$, rather than sampling them from $\mathcal{N}(0, C_{\sigma})$.

\paragraph{Setup}
We train \ourtok{} with perceptual loss~\cite{lpips,perceptualloss} and GAN loss~\cite{patchgan}, following \cite{sd,vqgan}.
We initialize the encoder from the base-size BEiT-3~\cite{beit3} checkpoint, and append a randomly initialized decoder.
Both encoder and decoder have 12 Transformer layers, totaling 172 million parameters.
The image patch size is 16.
We train tokenizers on the ImageNet training set~\cite{imagenet} with 200 epochs.
The batch size is 256.
The optimizer is AdamW~\cite{adamw} with $\beta=(0.0, 0.99)$ and a learning rate of 3e-4.
The weight decay is set to 0.01.
We apply layer-wise learning rate decay~\cite{beit} of 0.65 on the encoder.
For DiT and \our{} training, we follow the training recipes of \cite{dit}.
More training details are presented in \Cref{app:setup:image}.

\Cref{fig:var_fid} presents the FID-50K scores of DiT and \our{} using tokenizers with different variance.
The ``stars'' in the figure represent tokenizers that were tuned for previous latent diffusion models~\cite{sd}, which usually have a small variance, i.e., being more like an autoencoder instead of VAE.
The other ``dots'' are \ourtok{} with fixed variance.
We summarize the findings as follows:

\textbf{The tokenizers tuned for previous image-level diffusion models are ineffective for \our{}.}
For \our{}, the ``stars'' (in \Cref{fig:var_fid}) perform significantly worse than the others that have larger tokenizer variances.
The results indicate that directly adopting tokenizer configurations from previous diffusion models is suboptimal for \our{}.
The tokenizers with small variances are not robust to autoregressive error~\cite{givt}.

\textbf{\our{} favors tokenizers with larger variances.}
For the example without classifier-free guidance (i.e., CFG=1.0 in \Cref{fig:var_fid}), \our{} improves monotonically with increased variance.
In contrast, the choice of variance is not critical for DiT models.
The analysis highlights the advantage of \ourtok{}, whose variance is easily controllable.
So we recommend to use re-trained \ourtok{} as tokenizers for \our{}, rather than directly using previous ones.

\subsubsection{Inference Efficiency}
\label{sec:infer:efficiency}

\begin{figure}
    \centering
    \begin{subfigure}[b]{0.44\textwidth}
        \includegraphics[width=\linewidth]{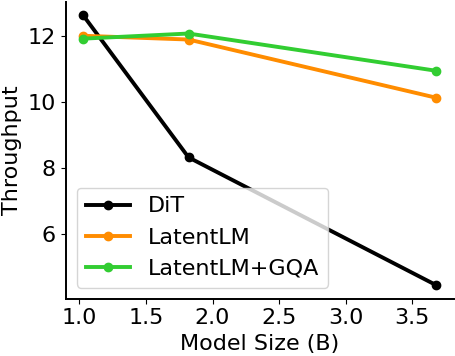}
        \caption{Throughput with increasing model sizes.}
        \label{fig:inference_bsz128}
    \end{subfigure}
    \hfill
    \begin{subfigure}[b]{0.44\textwidth}
        \includegraphics[width=\linewidth]{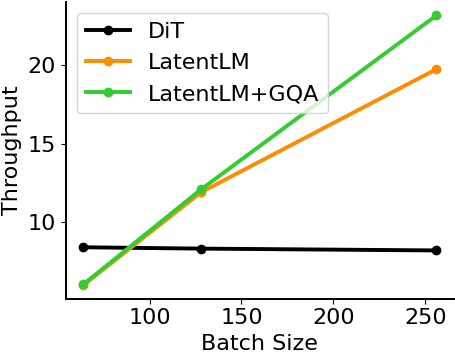}
        \caption{Throughput with increasing batch sizes.}
        \label{fig:inference_XL}
    \end{subfigure}
\caption{We compare the inference throughput of Diffusion Transformer (DiT)~\cite{dit} and \our{} in the settings of different model size and batch size. ``GQA'' stands for group-query attention~\cite{gqa}.}
\label{fig:image:efficiency}
\end{figure}

As shown in \Cref{fig:image:efficiency}, we investigate the inference capabilities of \our{} by examining the effects of model size and batch size.
We perform efficiency comparisons using 20 diffusion inference steps on a single H100 GPU.

First, we evaluate models ranging from 1B to 3.8B parameters with a fixed batch size of 128.
\Cref{fig:inference_bsz128} shows that DiT's throughput decreases significantly with larger model size.
Because DiT has to iteratively perform multiple forward passes, it incurs higher computational costs.
For the largest model with 3.8B parameters, \our{} achieves a 2.47$\times$ increase in throughput, demonstrating its scalability advantages.

As presented in \Cref{fig:inference_XL}, we then assess the 1.82B models with varying batch sizes.
As the batch size increases, the throughput of \our{} scales favorably with DiT.
In addition, group-query attention (GQA)~\citep{gqa} further improves throughput.
For a batch size of 256, our approach achieves a 2.84$\times$ throughput improvement.
The results indicate that \our{} benefits from significantly reduced FLOPs compared to image-level diffusion models, particularly at larger batch sizes.
Additional experiments on other model sizes are provided in \Cref{app:inference}.

\subsection{Multimodal LLMs: Unified Understanding and Generation}
\label{sec:exp:mllm}

We train multimodal large language models with \our{} for unified understanding and generation.
In this section, we focus on vision-language models.
By unifying next-token prediction and diffusion, the model can seamlessly handle interleaved image-text data, text-only data, and image-text pairs.
The proposed method simplifies the multimodal training and inference processes, allowing to learn in context (e.g., few-shot), follow multimodal instructions, and perform multimodal dialogue.
Moreover, unified modeling enables new capabilities.
For example, we can edit or generate images by conditioning on text and multiple input images.

\subsubsection{Training Setup}

\paragraph{Training Data}
We use three types of data in the training stage: text-only data, image-text pair data, and interleaved text-image data. The mix-up ratio is 2:1:1. The data sources are described as follows:
\begin{itemize}[leftmargin=*]
\setlength\itemsep{0.01em}
\item \textbf{Text-Only Data} The text training corpus follows~\cite{yoco}, including Common Crawl, RefinedWeb~\cite{RefinedWeb}, and StarCoder~\cite{starcoder}.
\item \textbf{Image-Text Pairs} We follow \cite{kosmos-1,kosmos-2} to construct the paired data, i.e., English LAION-2B~\cite{laion5b}, LAION-400M~\cite{laion400m}, COYO-700M~\cite{coyo700m}, and Conceptual Captions~\cite{cc3m,cc12m}.
\item \textbf{Interleaved Image-Text Data} We use the same interleaved multimodal documents as in \cite{kosmos-1,kosmos-2}. The web pages are filtered from Common Crawl archives. The documents are interleaved with text and image.
\end{itemize}

\paragraph{Configuration}
We train a 1.3B-size Transformer as the backbone. We set the hidden size to 2048. The number of layers is 24.
The training sequence length is 4096. We use \texttt{tiktoken-cl100k\_base} as the text tokenizer.
The batch size is 4M tokens.
We use the AdamW~\cite{adamw} optimizer with $\beta=(0.9,0.98)$.
The maximal learning rate is 3e-4 with 500 warmup steps.
The total schedule is set to 1T tokens.
We train the model with 50k steps (i.e., 200B tokens) for comparison.
More hyperparameters are detailed in \Cref{app:hp:mllm}.

\subsubsection{Results}

We compare \our{} with Transfusion~\cite{transfusion}, and vector quantized models (VQ-MLLM; i.e., the models using vector quantized image tokenizers).
Specifically, Transfusion shares Transformer weights for autoregressive language modeling and image-level diffusion, which uses bidirectional iterative denoising for images and causal masking for text.
Moreover, VQ-MLLM uses VQ-VAE~\cite{vqvae,vqgan} as the tokenizer for images, where images are compressed to discrete code.
We use the VQ-VAE tokenizer open-sourced by LlamaGen~\cite{llamagen} in VQ-MLLM.
We use the same training configuration and tokenizer settings for comparison.
To align the number of parameters, we use a 6-layer ViT as the image head of Transfusion.

\begin{table}[t]
\centering
\begin{tabular}{lccccc}
\toprule
\multirow{2}{*}{\textbf{Model}} & \textbf{Text} & \multicolumn{2}{c}{\textbf{Text-to-Image}} & \multicolumn{2}{c}{\textbf{Image-to-Text}} \\
& \bf Valid PPL$\downarrow$ & \bf FID$\downarrow$ & \bf CLIP$\uparrow$ & \bf MS-COCO$\uparrow$ & \bf VQAv2$\uparrow$ \\
\midrule
VQ-MLLM & 2.79 & 16.92 & \textbf{29.33} & 37.4 & 30.19 \\
Transfusion & 2.74 & 16.10 & 28.66 & 43.4 & 35.36 \\
\our{} & \textbf{2.73} & \textbf{14.54} & 28.75 & \textbf{54.5} & \textbf{38.72} \\
\bottomrule
\\
\end{tabular}
\caption{Results of multimodal large language models on text language modeling, image-to-text, and text-to-image generation.
We compare with Transfusion~\cite{transfusion} and vector quantized models (VQ-MLLM; i.e., using discrete code to represent images).
``PPL'' is perplexity. CLIP~\cite{clip} score measures the similarity. We report CIDEr~\cite{cider} score for MS-COCO~\cite{mscoco} and accuracy for VQAv2~\cite{vqav2}.}
\label{tbl:mllm}
\end{table}

\paragraph{Language Modeling}
\Cref{tbl:mllm} presents the evaluation results on language modeling, text-to-image generation, and multimodal understanding.
First, \our{} achieves a better perplexity in language modeling.
The results indicate that our method tends to better share knowledge between modalities with less conflicts.
The similarity between next-token prediction and next-token diffusion also benefits the unified modeling.

\begin{figure}[t]
\centering
\begin{subfigure}[b]{0.44\textwidth}
\includegraphics[width=\linewidth]{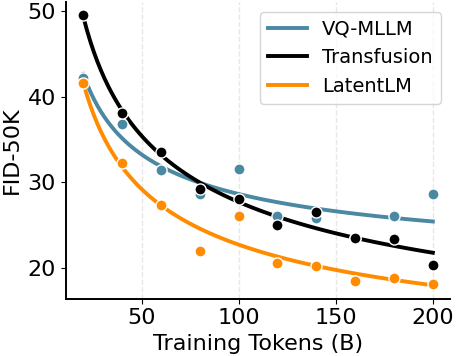}
\caption{Text-to-image FID~\cite{fid}.}
\label{fig:mllm_t2i_fid}
\end{subfigure}
\hfill
\begin{subfigure}[b]{0.44\textwidth}
\includegraphics[width=\linewidth]{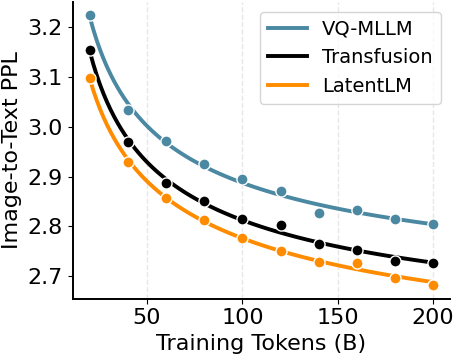}
\caption{Image-to-text validation perplexity.}
\label{fig:mllm_i2t_ppl}
\end{subfigure}
\caption{We scale up the number of training tokens for multimodal large language models. \our{} outperforms vector quantized models (VQ-MLLM) and Transfusion~\cite{transfusion} for both text-to-image and image-to-text generation. The FID scores are evaluated on MS-COCO~\cite{mscoco}.}
\end{figure}

\begin{figure}[t]
\centering
\begin{minipage}[t]{0.24\textwidth}
\centering
\includegraphics[width=\linewidth]{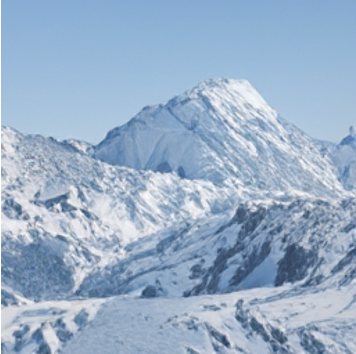}
\subcaption{A majestic mountain range covered in snow.}
\end{minipage}%
\hspace{0.25em}
\begin{minipage}[t]{0.24\textwidth}
\centering
\includegraphics[width=\linewidth]{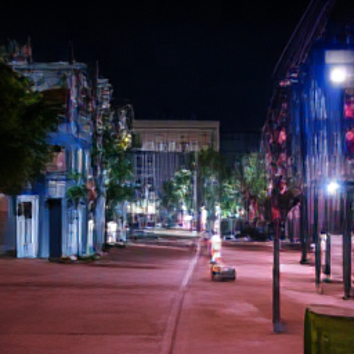}
\subcaption{A city street illuminated by lights.}
\end{minipage}%
\hspace{0.25em}
\begin{minipage}[t]{0.24\textwidth}
\centering
\includegraphics[width=\linewidth]{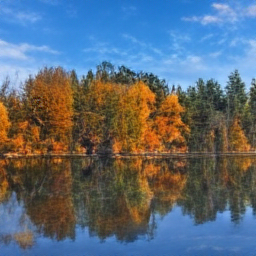}
\subcaption{A crystal lake surrounded by autumn trees.}
\end{minipage}%
\hspace{0.25em}
\begin{minipage}[t]{0.24\textwidth}
\centering
\includegraphics[width=\linewidth]{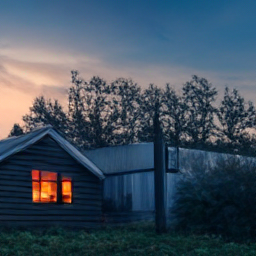}
\subcaption{A small house in a wooden at sunset.}
\end{minipage}
\caption{Text-to-image examples of \our{}.}
\label{fig:t2i_samples}
\end{figure}

\paragraph{Text-to-Image Generation}
Then we evaluate text-to-image generation on MS-COCO~\cite{coco}.
\Cref{tbl:mllm} shows that \our{} achieves lower FID scores, i.e., better generation quality.
The trend is also consistent with \Cref{tbl:imagenet_fid}, where Transfusion is aligned with DiT, and VQ-MLLM with LlamaGen.
In addition, \Cref{fig:mllm_t2i_fid} presents the scaling curves in terms of the number of training tokens, where \our{} consistently achieves better FID scores.
It is worth noting that the performance of VQ-MLLM seems saturated compared to the other methods.
\Cref{fig:t2i_samples} also shows several text-to-image samples of \our{}.

\paragraph{Image-to-Text Generation}
\Cref{tbl:mllm} reports image captioning on MS-COCO~\cite{coco} and visual question answering on VQAv2~\cite{vqav2}.
\our{} achieves better performance in both multimodal understanding tasks.
Compared to VQ-MLLM, the continuous representations used by Transfusion and \our{} are more lossless than discrete code.
Compared to Transfusion, \our{} keeps training and inference consistent, rather than adding noise to input images during training.
\Cref{fig:mllm_i2t_ppl} presents the text perplexity on the image-to-text validation data.
The results are also consistent with those reported in \Cref{tbl:mllm}.

\subsection{Text-to-Speech Synthesis: Higher Compression Ratio, Fewer Decoding Steps}
\label{sec:exp:tts}

We apply \our{} to text-to-speech synthesis (TTS).
Due to continuous representations, \ourtok{} achieves superior reconstruction results with a significantly higher compression ratio and lower frame rate than previous speech tokenizers~\cite{encodec,dac,dac_lowbit,wavtokenizer,kyutai2024moshi}.
\our{} outperforms the state-of-the-art VALL-E 2~\cite{valle2} model on both speaker similarity score and robustness while requiring 10$\times$ fewer decoding steps.

\subsubsection{Training Setup}

Considering the variable-length nature of speech data, our speech tokenizer employs a convolutional architecture that supports streaming encoding and decoding.
Specifically, \ourtok{} for speech consists of a convolutional encoder, a continuous VAE quantizer, and a convolutional decoder.
The encoder comprises multiple stages and downsampling layers organized in a hierarchical structure.
Each stage includes several ConvNeXt blocks~\cite{convnext}, where the original 2D convolution is replaced with 1D causal convolution.
For compression ratios of 1600, 3200, and 6400, the downsampling layer reduces the input waveform by factors of [2, 4, 5, 5, 8], [4, 4, 5, 5, 8], and [4, 5, 5, 8, 8], respectively.
Each time the downsampling layer is applied, the number of channels doubles, starting from 32 and increasing to 1024.
The encoder contains around 120 million parameters in total.
The decoder is a mirror of the encoder. 
As for the discriminator, we use the multi-period discriminator~\cite{hifigan} and the complex STFT discriminator in DAC~\cite{dac}.

The hidden size of \our{} is 1024, with 24 layers and 16 attention heads.
The intermediate FFN dimension is set to 4096.
The diffusion head contains three layers of feedforward networks.
We use the same Transformer architecture as VALL-E 2~\cite{valle2} for comparison.
Additional hyperparameters are described in \Cref{app:hp:tts}.

\subsubsection{Training Data}

\paragraph{Tokenizer}
We train \ourtok{} on a large and diverse corpus that includes speech, audio, and music. 
For speech, we use the clean speech subset from DNS Challenge 4~\cite{dns} and all splits from the Common Voice v7 dataset~\cite{commonvoice:2020}. 
For audio, we use the FSD50K dataset~\cite{fonseca2021fsd50k}, along with the balanced and unbalanced splits from AudioSet~\cite{audioset}. 
For music, we use the MUSDB dataset~\cite{rafii2017musdb18} and the Jamendo dataset~\cite{jamendo}.
All the data are resampled to 24kHz monophonic format.

\paragraph{TTS Model}
We utilize Libriheavy corpus~\cite{libriheavy} as training data following VALL-E 2~\cite{valle2}.
This corpus is a labeled version of the Librilight corpus~\cite{librilight}, which features 50,000 hours of speech from approximately 7,000 different speakers, sourced from open-access English audiobooks associated with the LibriVox project\footnote{\url{https://librivox.org}}.

\subsubsection{Evaluation Metrics}

We evaluate our speech tokenizer using several automatic metrics, including: \textbf{Mel Distance}, which measures the distance between log Mel spectrograms as configured in DAC~\cite{dac}; \textbf{PESQ-WB}~\cite{pesq}, an intrusive metric for speech quality by comparing perceptual differences; \textbf{STOI}~\cite{stoi}, which assesses speech intelligibility through short-time segment correlation; \textbf{VISQOL}~\cite{visqol}, a perceptual quality metric based on spectral similarity; \textbf{UTMOS}~\cite{utmos}, a reference-free mean opinion score for audio quality; \textbf{Speaker Similarity} (\textbf{SIM}), measured using WavLM-TDNN~\cite{chen2022wavlm}; and \textbf{Word Error Rate} (\textbf{WER}), calculated using both Conformer-Transducer~\cite{gulati2020conformer} (WER-C) and HuBERT-Large~\cite{hubert} (WER-H) models.

\subsubsection{System Evaluation}
\label{sec:exp:tts:results}

\begin{table}[t]
\centering
\resizebox{\textwidth}{!}{
\begin{tabular}{lccccccccc}
\toprule
\multirow{2}{*}{\bf System} & \multirow{2}{*}{\tabincell{c}{\bf Frame Rate \\ \bf Length/s}$\downarrow$} && \multicolumn{3}{c}{\bf Ref Utterance as Prompt} && \multicolumn{3}{c}{\bf 3s Prefix as Prompt} \\  
\cmidrule{4-6} \cmidrule{8-10} &&& \bf SIM$\uparrow$ & \bf WER-C$\downarrow$ & \bf WER-H$\downarrow$ && \bf SIM$\uparrow$ & \bf WER-C$\downarrow$ & \bf WER-H$\downarrow$ \\ 
\midrule
Ground Truth & - && 0.779 & 1.6 & 2.2 && 0.668 & 1.6 & 2.2 \\
\midrule
VALL-E 2~\cite{valle2} & 75 && 0.643 & 1.5 & 2.4 && 0.504 & 1.6 & 2.3 \\
Voicebox~\cite{voicebox} & 100 && 0.662 & - & 1.9 && \bf 0.593 & - & \bf 2.0 \\
MELLE~\cite{melle} & 62 && 0.625 & 1.5 & 2.1 && 0.508 & 1.5 & \bf 2.0 \\
\midrule
\our{} & 15 && \bf 0.697 & \bf 1.2 & 1.8 && 0.571 & \bf 1.4 & \bf 2.0 \\
\our{} & 7.5 && 0.656 & \bf 1.2 & \bf 1.7 && 0.532 & 1.6 & 2.3 \\
\our{} & 3.75 && 0.598 & 1.7 & 2.3 && 0.467 & 3.1 & 4.5 \\
\bottomrule
\\
\end{tabular}
}
\caption{\our{} outperforms previous systems on zero-shot speech synthesis in both settings. Moreover, the number of decoding steps is much less than others, achieving faster inference speed. The results are reported on LibriSpeech test-clean set. The WER-H and SIM results of VALL-E 2 using 3s prefix as prompt are from \cite{melle}.}
\label{tbl:tts_librispeech}
\end{table}

Table~\ref{tbl:tts_librispeech} presents zero-shot text-to-speech (TTS) results on the LibriSpeech test-clean dataset. We evaluate the synthesis quality under two distinct settings: (1) using a reference utterance from the same speaker as the prompt, and (2) evaluating speech continuation by using the first 3 seconds of speech as the prompt.

Our model, operating at a frame rate of 15 (i.e., generating 1 second of speech in 15 autoregressive steps), surpasses previous state-of-the-art methods when using a same-speaker reference utterance as the prompt. \our{} with a frame rate of 7.5 achieves superior performance compared to the neural codec language model VALL-E 2~\cite{valle2}, while requiring an order of magnitude ($10\times$) fewer autoregressive inference steps. Moreover, \our{} eliminates the need for the non-autoregressive (NAR) model employed in VALL-E 2, resulting in improved computational efficiency. Even at a lower frame rate of 3.75, \our{} maintains competitive performance.
The higher compression ratio reduces the sequence length, which in turn greatly accelerates the decoding speed.

\subsubsection{Evaluating the Quality of Tokenizers}
\label{sec:exp:tts:tokenizer}

\begin{table}[t]
\centering
\resizebox{\textwidth}{!}{
\begin{tabular}{lccc|ccccc}
\toprule
\bf Tokenizer & \bf $\mathbf{N_q}$$\downarrow$ & \bf \tabincell{c}{Frame \\ Rate}$\downarrow$ & \bf \tabincell{c}{Comp. \\ Ratio}$\uparrow$ & \bf \tabincell{c}{Mel \\ Dist.}$\downarrow$ & \bf PESQ$\uparrow$ & \bf STOI$\uparrow$ & \bf VISQOL$\uparrow$ & \bf UTMOS$\uparrow$ \\
\midrule
\multicolumn{9}{l}{~~\textit{Tokenizers with lower compression ratio}} \\
Encodec~\cite{encodec} & 32 & 75 & 10 & 0.823 & 3.591 & 0.962 & 4.536 & 3.195 \\
DAC~\cite{dac} & 32 & 75 & 10 & \bf 0.355 & \bf 4.424 & \bf 0.987 & \bf 4.914 & \bf 3.469 \\
Encodec~\cite{encodec} & 8 & 75 & 40 & 0.987 & 2.687 & 0.925 & 4.258 & 2.656 \\
DAC~\cite{dac} &  8 & 75 &  40 & 0.707 & 3.329 & 0.941 & 4.485 & 3.133 \\
DAC$_\text{low}$~\cite{dac_lowbit} & 4 & 75 &  80 & 0.753 & 3.107 & 0.938 & 4.391 & 3.453 \\
DAC$_\text{low}$~\cite{dac_lowbit} & 2 & 75 &  160 & 0.916 & 2.269 & 0.896 & 3.981 & 3.297 \\
Mimi~\cite{kyutai2024moshi} & 8 & 12.5 & 240 & 0.987 & 3.217 & 0.946 & 4.332 & 3.375 \\
\midrule
\multicolumn{9}{l}{~~\textit{Tokenizers with higher compression ratio}} \\
WavTokenizer~\cite{wavtokenizer} & 1 & 75 & 320 & 0.871 & 2.266 & 0.891 & 4.120 & 3.432 \\
Mimi~\cite{kyutai2024moshi} & 4 & 12.5 & 480 & 1.458 & 1.568 & 0.826 & 3.390 & 2.652 \\
WavTokenizer~\cite{wavtokenizer} & 1 & 40 & 600 & 1.037 & 1.670 & 0.834 & 3.782 & 3.053 \\
\ourtok{}$_{32}$ & 1 & 15 & 1600 & 0.813 & 2.724 & 0.926 & 4.268 & 3.491 \\
\ourtok{}$_{64}$ & 1 & 7.5 & 3200 & \bf 0.798 & \bf 2.756 & \bf 0.929 & \bf 4.289 & \bf 3.505 \\
\ourtok{}$_{128}$ & 1 & 3.75 & 6400 & 0.852 & 2.533 & 0.916 & 4.165 & 3.460 \\
\bottomrule
\\
\end{tabular}
}
\caption{The \ourtok{} tokenizers obtain competitive reconstruction quality while having high compression ratio.
We report results on the LibriTTS test-other set.
``$N_q$'' represents the number of quantizers.
We define the compression ratio as the audio sample rate divided by $N_q$ and the frame rate.
``\ourtok{}$_{32}$'' denotes that the latent dimension of the tokenizer is 32.}
\label{tbl:sp_tok}
\end{table}

\Cref{tbl:sp_tok} compares \ourtok{} and other codec models on the LibriTTS test-other set.
\ourtok{} achieves better reconstruction quality in a compression ratio of 1600$\times$ compared to Encodec~\cite{encodec} (40$\times$), DAC$_\text{low}$~\cite{dac_lowbit} (160$\times$), WavTokenizer~\cite{wavtokenizer} (320$\times$), and Mimi~\cite{kyutai2024moshi} (480$\times$).
Notably, as we further increase the compression ratio, the reconstruction quality does not deteriorate significantly.
At a compression ratio of 6400, the resulting sequence length when used in a language model is already comparable to BPE tokenization~\cite{bpe}, approaching a 1:1 ratio.

\subsubsection{Ablation Studies}

\begin{table}[t]
\centering
\resizebox{\textwidth}{!}{
\begin{tabular}{ccc|ccccccc}
\toprule
\multirow{2}{*}{\tabincell{c}{\bf Compression \\ \bf Ratio}}  & \multirow{2}{*}{\tabincell{c}{\bf Frame \\ \bf Rate}} & \multirow{2}{*}{\tabincell{c}{\bf Latent \\ \bf Dimension}} && \multicolumn{3}{c}{\bf \ourtok{} Reconstruction} && \multicolumn{2}{c}{\bf Zero-Shot TTS} \\  
\cmidrule{5-7} \cmidrule{9-10}  &&&& \bf Mel Dist.$\downarrow$ & \bf SIM$\uparrow$ & \bf WER-C$\downarrow$ && \bf SIM$\uparrow$ & \bf WER-C$\downarrow$ \\ 
\midrule
640$\times$ & 37.5 & 16 && 0.929 & 0.866 & 1.9 && 0.655 & 1.4 \\
1600$\times$ & 15 & 16 && 1.080 & 0.700 & 2.7 && 0.545 & 1.6 \\
1600$\times$ & 15 & 32 && 0.950 & 0.870 & 1.9 && 0.661 & 1.5 \\
\bottomrule
\\
\end{tabular}
}
\caption{Ablation results of different \ourtok{} compression ratios and latent dimensions. We report tokenizer reconstruction quality and zero-shot speech synthesis.}
\label{tbl:abla:latent_dim}
\end{table}

\paragraph{Compression Ratio and Latent Dimension}
We find that increasing the latent dimension enables the model to achieve a higher compression ratio and a lower frame rate.
Table~\ref{tbl:abla:latent_dim} presents the \ourtok{} reconstruction and zero-shot speech synthesis results with different compression ratios and latent dimensions.
We report the in-domain Mel distance performance of \ourtok{}, along with the speaker similarity score and WER-C for tokenizer reconstruction and zero-shot speech generation on the LibriSpeech test-clean set.
We use a 12-layer Transformer model for the TTS ablation studies.
If the latent dimension remains unchanged, a higher compression ratio leads to a decrease in reconstruction performance and TTS speaker similarity score. 
However, by increasing the latent dimension of \ourtok{}, we can compensate for this loss, allowing our model to use a higher compression ratio and a lower frame rate.
Our model can generate 1 second of speech using significantly fewer autoregressive inference steps, compared to VALL-E 2.

\begin{figure}[t]
\centering
\begin{subfigure}[ht]{0.49\textwidth}
\includegraphics[width=\linewidth]{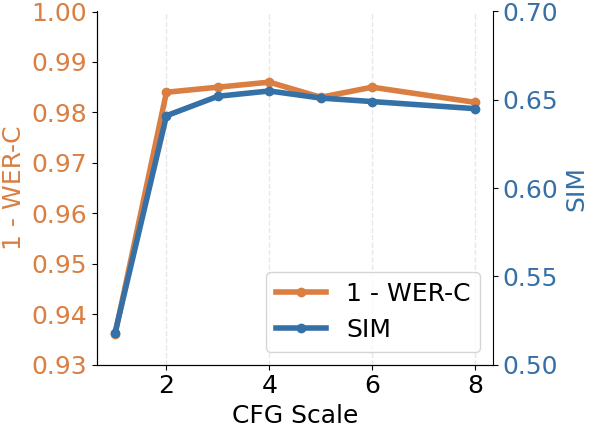}
\caption{Results using different CFG scales.}
\label{fig:tts_cfg_scales}
\end{subfigure}
\hfill
\begin{subfigure}[ht]{0.49\textwidth}
\includegraphics[width=\linewidth]{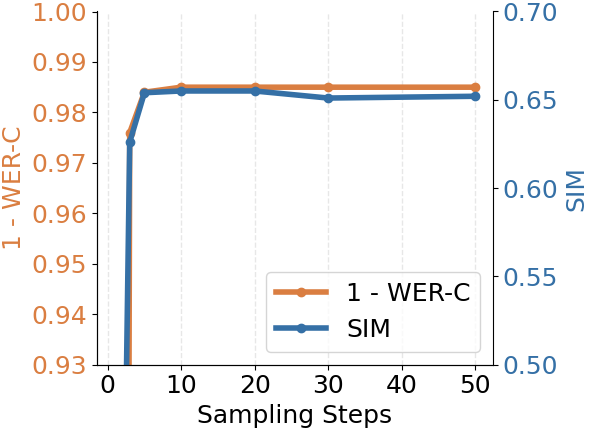}
\caption{Results using different sampling steps.}
\label{fig:tts_sampling_steps}
\end{subfigure}
\caption{Ablation results of different CFG~\cite{cfg} scales and inference sampling steps. We report zero-shot speech synthesis results.}
\end{figure}

\paragraph{CFG Scale}
Figure~\ref{fig:tts_cfg_scales} illustrates the zero-shot speech synthesis results using classifier-free guidance (CFG)~\cite{cfg}. 
When the CFG scale is set to 1, CFG is not applied. 
The use of classifier-free guidance significantly enhances the model's performance. 
Furthermore, we find that setting the CFG scale to 4 yields the best results.

\paragraph{Inference Sampling Step}
Figure~\ref{fig:tts_sampling_steps} presents the results of zero-shot speech synthesis using different inference sampling steps of the diffusion head. 
We set the CFG scale to 4 for the ablation studies. 
More sampling steps require more inference time. 
We find that a sampling step of 3 yields competitive results, and increasing it to 5 leads to further improvement. 
When the sampling step is increased further, the results improve only slightly.
Using a sampling step of 5 allows the model to achieve strong performance while maintaining a fast inference speed.

\section{Conclusion and Future Work}

The work can be advanced from the following perspectives:
\begin{itemize}[leftmargin=*]
\setlength\itemsep{0.01em}
\item \textbf{Latent Multimodal Reasoning} The proposed unified modeling facilitates complex multimodal reasoning tasks that require simultaneous understanding of multiple modalities. For instance, self-reflection can automatically correct produced images, which requires the multimodal language model to understand the generated image without encoding it again. Moreover, multimodal-native reasoning enables the model to track the search states via latent vectors, for example, step-by-step plotting the planned trajectory on the image of input map.
\item \textbf{Video Generation and World Modeling} The autoregressive nature of \our{} fits well with video data, which shows particular promise in maintaining temporal consistency and spatial coherence. Moreover, \our{} can perform planning by generating scripts and videos in an interleaved way, making it particularly suitable for long-video generation. The approach's higher compression ratio compared to traditional quantization methods enables efficient generation without sacrificing visual quality. In addition, we can integrate actions to control and simulate the interactive environment, which can be used as a world model.
\item \textbf{Cross-Modal Transfer Between Speech and Text} Because of the high compression ratio of continuous representations, we can use similar tokenization granularity for speech and text data, which tends to ease knowledge transfer across modalities. Similarly, multilingual pretraining enables zero-shot cross-lingual transfer, where training on English benefits other languages. It is useful to achieve seamless code switch between speech and text and opens new opportunities for user interface.
\item \textbf{Embodied AI and Robot Action} By representing robot actions as continuous data, we enable end-to-end learning of robot behaviors that can be seamlessly integrated with language instructions, visual observations, and other sensory inputs. The unified framework simplifies the development of robots that can understand commands in natural language, learn from demonstrations, and adapt to new environments while maintaining a consistent internal representation across all modalities.
\item \textbf{Text Data} We can also apply latent language modeling to text data, rather than predicting words as discrete tokens. The VAE tokenizer tends to achieve a higher compression ratio than previous discrete tokenizers. The shorter sequence length improves the generation efficiency by reducing the autoregressive steps.
\end{itemize}

\section*{Acknowledgement}

We would like to acknowledge Ben Huntley for maintaining the GPU cluster, and Zhikang Niu for the help of training speech tokenizer.
We implement DiT~\cite{dit} based on \url{https://github.com/facebookresearch/DiT}.
The experiments on multimodal large language models utilize the curated data from \textsc{Kosmos}~\cite{kosmos-1,kosmos-2} and \textsc{RedStone}~\cite{redstone}. The implementation is based on the TorchScale~\cite{torchscale} library.

\nocite{givt}
\nocite{llamagen}
\nocite{jetformer}
\nocite{mar}

\bibliographystyle{alpha}
\bibliography{latentlm}

\newpage
\appendix

\section{Hyperparameters for Image Generation Scaling}
\label{app:hp:scaling}

\Cref{tbl:hp:scaling} details the hyperparameters used for \Cref{sec:exp:scaling}, where we compare the scalability properties of Diffusion Transformer (DiT)~\cite{dit} and \our{}.
We describe the hidden dimension, the number of layers, and the number of heads for the models.
Specifically, we follow \cite{dit} for the DiT architecture.
In addition, we augment DiT with RMSNorm~\cite{rmsnorm} and SwiGLU~\cite{swish,glu}.
To align the number of parameters, the FFN size for DiT is set to $\frac{8}{3}d$, while for \our{}, it is set to $4d$.
We train the models for 75,000 steps, which corresponds to approximately 120 epochs, to facilitate scaling comparisons.

\begin{table}[ht]
\centering
\begin{tabular}{lcccccc}
\toprule
& \bf Size & \bf Hidden Dim. & \bf \#Layers  & \bf \#Heads & \bf Learning Rate \\
\midrule
Medium & 455M & 1024 & 24 & 16 & $8\times10^{-4}$ \\
Large & 1.03B & 1536 & 24 & 12 & $3\times10^{-4}$ \\
XL & 1.82B & 2048 & 24 & 16 & $2\times10^{-4}$ \\
3B & 3.68B & 2560 & 32 & 20 & $1.6\times10^{-4}$ \\
\bottomrule
\\
\end{tabular}
\caption{Model size and hyperparameters used for the scaling experiments in~\Cref{sec:exp:scaling}.}
\label{tbl:hp:scaling}
\end{table}

\section{Hyperparameters for Tokenizer Analysis}
\label{app:setup:image}

We present the hyperparameters used for \Cref{sec:exp:tok}.
We follow the training recipes of \cite{dit} for DiT and \our{} training.
We set the hidden size to 1024. The number of layers is 24.
Because \our{} does not have AdaLN in the Transformer backbone, we adjust the intermediate FFN dimension (i.e., 2730 in DiT, and 4096 in \our{}) to match their model size.
The diffusion head has three layers of feedforward networks.

We use the AdamW~\cite{adamw} optimizer with $\beta=(0.9,0.98)$.
We use the cosine learning rate schedule with a maximal value of 1e-4 and 100 warmup steps.
The weight decay is $0.1$.
We train models using a batch size of 256 for 200,000 steps, which is approximately equivalent to 40 epochs.
We use the cosine beta schedule and v-prediction~\cite{v_pred} for diffusion. 
We use DDPM~\cite{ddpm} with 1000 steps during training.
DPM-Solver~\cite{dpm-solver,dpm-solver++} with 20 steps is used during inference.

\section{Inference Efficiency with Different Model Sizes}
\label{app:inference}

\begin{figure}[h]
    \centering
    \begin{subfigure}[b]{0.24\textwidth}
        \includegraphics[width=\linewidth]{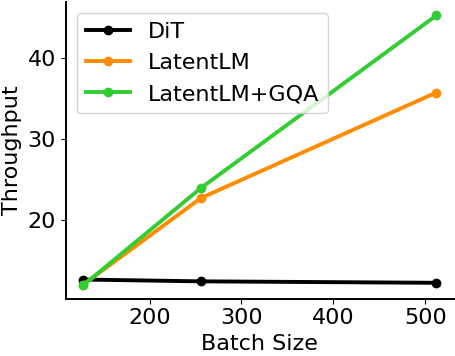}
        \caption{Model Size: 1.03B}
    \end{subfigure}
    \hfill
    \begin{subfigure}[b]{0.24\textwidth}
        \includegraphics[width=\linewidth]{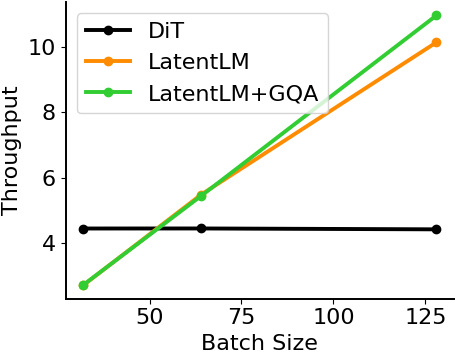}
        \caption{Model Size: 3.68B}
    \end{subfigure}
    \hfill
    \begin{subfigure}[b]{0.24\textwidth}
        \includegraphics[width=\linewidth]{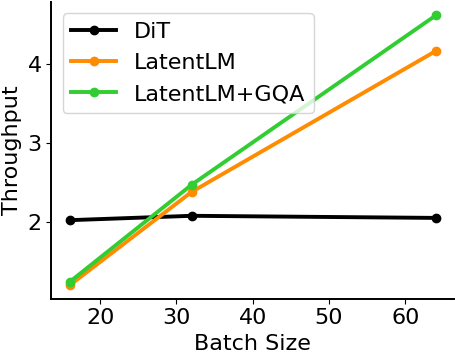}
        \caption{Model Size: 9.35B}
    \end{subfigure}
    \hfill
    \begin{subfigure}[b]{0.24\textwidth}
        \includegraphics[width=\linewidth]{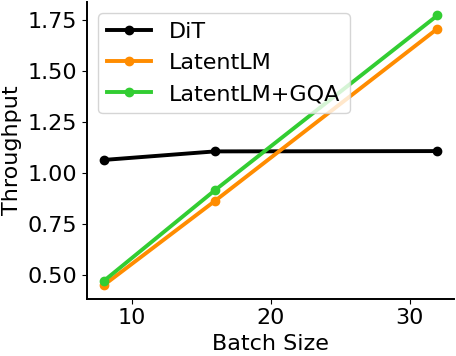}
        \caption{Model Size: 17.96B}
    \end{subfigure}
    \caption{Inference throughput of various model size and batch size. ``GQA'' stands for group-query attention~\cite{gqa}.}
    \label{fig:app:inference:efficiency}
\end{figure}

In \Cref{sec:infer:efficiency}, we compare the inference throughput for DiT and \our{} on a H100 GPU card.
As shown in \Cref{fig:app:inference:efficiency}, we evaluate the efficiency with various model size and batch size.
The results show that \our{}'s throughput increases with a larger batch size.
Our approach benefits from key-value caches of causal Transformers, which avoids recomputation of history predictions.
In contrast, DiT's throughput remains similar.
In addition, group-query attention (GQA)~\cite{gqa} further improves the inference efficiency of \our{}.
Another advantage is that we can directly reuse the inference infrastructure of large language models to deploy \our{}.

\section{Hyperparameters for Multimodal Large Language Model}
\label{app:hp:mllm}

\Cref{tbl:hp:mllm} details the hyperparameters employed for multimodal large language models, as described in \Cref{sec:exp:mllm}.

\begin{table}[h]
\centering
\begin{tabular}{lc}
\toprule
\textbf{Params} & \textbf{Values} \\
\midrule
Layers  & {24} \\
Hidden size & {2048} \\
FFN size & {6144} \\
Vocab size & 100,288 \\
Heads & {16} \\
Adam $\beta$ & {(0.9, 0.98)} \\
LR & $3\times10^{-4}$ \\
Batch size & {4M} \\
Warmup steps & {500} \\
Weight decay & {0.1} \\
Head Layers & {6} \\
\bottomrule
\\
\end{tabular}
\caption{Hyperparameters used for multimodal large language models in~\Cref{sec:exp:mllm}.
}
\label{tbl:hp:mllm}
\end{table}

\section{Hyperparameters for Text-to-Speech Synthesis}
\label{app:hp:tts}

\Cref{tbl:hp:tts} lists the hyperparameters utilized for multimodal large language models, as discussed in \Cref{sec:exp:tts}.

\begin{table}[h]
\centering
\begin{tabular}{lc}
\toprule
\textbf{Params} & \textbf{Values} \\
\midrule
Layers  & {24} \\
Hidden size & {1024} \\
FFN size & {4096} \\
Heads & {16} \\
Adam $\beta$ & {(0.9, 0.98)} \\
LR & $7.5\times10^{-4}$ \\
LR schedule & cosine \\
Batch size & {5M} \\
Warmup steps & {10k} \\
Training steps & {100k} \\
Weight decay & {0.01} \\
Head Layers & {3} \\
\bottomrule
\\
\end{tabular}
\caption{Hyperparameters used for text-to-speech synthesis in \Cref{sec:exp:tts}.
}
\label{tbl:hp:tts}
\end{table}

\end{document}

%% file: settings.tex
\usepackage{multirow}
\usepackage{amsmath}
\usepackage{capt-of}
\usepackage{tabularx}
\usepackage{epsfig}
\usepackage{amssymb}
\usepackage{amsfonts}
\usepackage{booktabs}
\usepackage{scalerel}
\usepackage[inline]{enumitem}
\usepackage{listings}
\usepackage{varwidth}
\usepackage[export]{adjustbox}
\usepackage{tikz}
\usetikzlibrary{tikzmark}

\usepackage{stmaryrd}
\usepackage{bbm}
\usepackage{wrapfig}
\usepackage{pifont}
\usepackage[noabbrev]{cleveref}

\newcommand{\tabincell}[2]{\begin{tabular}{@{}#1@{}}#2\end{tabular}}

\definecolor{deepblue}{rgb}{0,0,0.5}
\definecolor{officeblue}{RGB}{0,102,204}
\definecolor{deepred}{rgb}{0.6,0,0}
\definecolor{deepgreen}{rgb}{0,0.5,0}
\definecolor{mybrickred}{RGB}{182,50,28}

\definecolor{fillcolor}{RGB}{216,217,252}

\usepackage{etoolbox}
\usepackage{framed}

\newif\ifxetexorluatex
\ifxetex
  \xetexorluatextrue
\else
  \ifluatex
    \xetexorluatextrue
  \else
    \xetexorluatexfalse
  \fi
\fi
\newcommand*\quotesize{60} 
\newcommand*{\openquote}
   {\tikz[remember picture,overlay,xshift=-4ex,yshift=-2.5ex]
   \node (OQ) {\fontsize{\quotesize}{\quotesize}\selectfont``};\kern0pt}

\newcommand*{\closequote}[1]
  {\tikz[remember picture,overlay,xshift=4ex,yshift={#1}]
   \node (CQ) {\fontsize{\quotesize}{\quotesize}\selectfont''};}

\colorlet{shadecolor}{white}

\newcommand*\shadedauthorformat{\emph} 

\newcommand*\authoralign[1]{%
  \if#1l
    \def\authorfill{}\def\quotefill{\hfill}
  \else
    \if#1r
      \def\authorfill{\hfill}\def\quotefill{}
    \else
      \if#1c
        \gdef\authorfill{\hfill}\def\quotefill{\hfill}
      \else\typeout{Invalid option}
      \fi
    \fi
  \fi}
{\authoralign{#1}
\ifblank{#2}
   {\def\shadequoteauthor{}\def\yshift{-2ex}\def\quotefill{\hfill}}
   {\def\shadequoteauthor{\par\authorfill\shadedauthorformat{#2}}\def\yshift{2ex}}
\begin{snugshade}\begin{quote}\openquote}
{\shadequoteauthor\quotefill\closequote{\yshift}\end{quote}\end{snugshade}}

\newfloat{Code}{htbp}{Code}

%% file: math_commands.tex
\usepackage{amsmath,amsfonts,bm}

\def\eqref#1{equation~(\ref{#1})}
\def\Eqref#1{Equation~(\ref{#1})}

\def\1{\bm{1}}

\def\vh{{\bm{h}}}

\def\vx{{\bm{x}}}

\def\mI{{\bm{I}}}

\DeclareMathAlphabet{\mathsfit}{\encodingdefault}{\sfdefault}{m}{sl}
\SetMathAlphabet{\mathsfit}{bold}{\encodingdefault}{\sfdefault}{bx}{n}

\newcommand{\softmax}{\mathrm{softmax}}

